\documentclass{article}

\PassOptionsToPackage{numbers}{natbib}
     \usepackage[preprint]{neurips_2021}

\usepackage[utf8]{inputenc} % allow utf-8 input
\usepackage[T1]{fontenc}    % use 8-bit T1 fonts
\usepackage{hyperref}       % hyperlinks
\usepackage{url}            % simple URL typesetting
\usepackage{booktabs}       % professional-quality tables
\usepackage{amsfonts}       % blackboard math symbols
\usepackage{nicefrac}       % compact symbols for 1/2, etc.
\usepackage{microtype}      % microtypography
\usepackage{xcolor}         % colors
\usepackage{multicol}
\usepackage{lipsum}
\usepackage{mwe}
\usepackage{subfig}
\usepackage{mathtools}
\usepackage{algorithm}
\usepackage{algpseudocode}
\usepackage{placeins}
\usepackage{amsmath}
\usepackage[nameinlink,noabbrev]{cleveref}

\usepackage[title]{appendix}

\newtheorem{theorem}{Theorem}
\newtheorem{lemma}{Lemma}

\title{Reducing the Long Tail Losses in Scientific Emulations with Active Learning}
\author{%
  Y. H. Lim \\
  Machine Discovery Ltd.\\
  Oxford, United Kingdom \\
  \texttt{yi.heng@machine-discovery.com} \\
  \AND
  M. F. Kasim \\
  Machine Discovery Ltd. \\
  Oxford, United Kingdom \\
  \texttt{muhammad@machine-discovery.com} \\
}

\begin{document}

\maketitle

\begin{abstract}
Deep-learning-based models are increasingly used to emulate scientific simulations to accelerate scientific research. However, accurate, supervised deep learning models require huge amount of labelled data, and that often becomes the bottleneck in employing neural networks. In this work, we leveraged an active learning approach called core-set selection to actively select data, per a pre-defined budget, to be labelled for training. To further improve the model performance and reduce the training costs, we also warm started the training using a shrink-and-perturb trick. We tested on two case studies in different fields, namely galaxy halo occupation distribution modelling in astrophysics and x-ray emission spectroscopy in plasma physics, and the results are promising: we achieved competitive overall performance compared to using a random sampling baseline, and more \textit{importantly}, successfully reduced the larger absolute losses, i.e. the long tail in the loss distribution, at \textit{virtually no overhead costs}.
\end{abstract}

\section{Introduction}
Simulators used in scientific experiments are often accurate but computationally expensive. To mitigate the problem, deep-learning-based emulators can be built to emulate the simulators \citep{brockherde2017bypassing, kwan2015cosmic, peterson2017zonal, rupp2012fast}. However, supervised neural networks require a large number of labelled training data, which are expensive to collect from simulations. Given the limited labelled data that can be generated, the quality of the data and the chosen deep learning model become ever more important. One way to address the latter is by employing efficient neural architecture search \citep{kasim2021building} to find the architecture that is tailored to the problem on a case-by-case basis, although the training could be expensive. 

In this work, we turn our attention to address the former, i.e. choosing good quality data. We hypothesise that, with a budget constraint, by selecting good quality data,
% by selecting data that are collectively representative of the full, original data distribution, 
not only could we obtain an emulator with a small mean error when tested, but also \textit{more importantly}, reduce the long tail of the loss distribution. To that end, we employ a diversity-based active learning approach, known as core-set selection \citep{sener2017active}, to iteratively augment labelled data to our training. We choose a single, deterministic, supervised CNNs-based model architecture for our emulator to reduce the training costs, but we note that the model can be further optimised in the future. To the best of our knowledge, previous works on active learning mostly focused on improving \textit{classification} accuracy \citep{sener2017active, gal2017deep, ash2019deep, kirsch2019batchbald}, and as such the loss distributions were not explicitly studied.

Our work applies active learning algorithm to build emulators for 2 scientific simulation cases using open-source data \citep{kasim2021building}: (1) x-ray emission spectroscopy (XES) for a pellet in inertial confinement fusion \citep{regan2013hot, ciricosta2017simultaneous} and (2) galaxy halo occupation distribution modelling (Halo) \citep{wake2011galaxy}.
The first simulation case (XES) takes 10 parameters describing the pellet design in inertial confinement fusion and produces a spectrum of the time-integrated x-ray emission during the fusion process.
The second case (Halo) takes 5 input parameters to produce angular-scale correlation of galaxy population.

Our contributions in this paper are two-fold:
\begin{enumerate}
    %\item We give a theoretical proof that the core-set approach \citep{sener2017active} can be extended to regression settings, and show both theoretically and empirically that the long tail losses in emulations can be reduced by selecting training samples using this core-set approach, if appropriate features are used.
    \item We give a theoretical proof that the core-set approach \citep{sener2017active} can be extended to regression settings, and show empirically that the long tail losses in emulations can be reduced by selecting training samples using this core-set approach 
    %\footnote{Full code and data have been made public, see: \hyperlink{https://github.com/machine-discovery/research}{https://github.com/machine-discovery/research}}.
    \item We further improve the emulator performance and reduce the overhead costs of the core-set approach to virtually none by using the shrink-and-perturb trick \citep{ash2019warm} and solving the k-center problem using KeOps library \citep{JMLR:v22:20-275}. Code and data are available \href{https://github.com/machine-discovery/research/tree/master/ALSciEmulation}{here}.
\end{enumerate}
% Active learning has recently drawn attention in the deep learning community, but the focus has largely been on classification problems. However, in scientific emulations, a lot of the problems are defined in the regression settings. We studied different methods in detail, including \citep{gal2017deep}, \citep{yoo2019learning} and \citep{sener2017active}, then adapted algorithms that are useful to the regression settings, giving empirical analyses in this paper.

\iffalse
\citep{gal2017deep} applied dropout at both training and test times to obtain epistemic uncertainty which is in turn used to select data that the model is most unconfident about. \citep{yoo2019learning} noted that the uncertainty-based method can be computationally demanding, hence proposed to attach a loss prediction module to the target model to predict the losses of the unlabelled data. \citep{sener2017active} argued
\fi

% \section{Prerequisites}
\section{Methods}
\subsection{Active Learning}
We consider an initial pool of unlabelled, identically and independently distributed ($i.i.d.$) data, $U = \{\mathbf{x}_{1}, \dots, \mathbf{x}_{n}\}$, where $\mathbf{x}_{i}$'s are the input feature vectors. In each active learning iteration $t \in \{1, \dots, T \}$, we sample $b_t$ data-points from the pool $U$ without replacement, query an oracle to obtain their corresponding $\mathbf{y}_i$'s, and update the model. In our case, the oracles are the respective scientific simulations. The labelled data collected in each time step $t$ are collectively denoted by $\mathbf{s}^t$, and the set of labelled data available up to $t$ is denoted $\{\mathbf{x}_{\mathbf{s}(j)}, \mathbf{y}_{\mathbf{s}(j)}\}_{j\in[m]}$, where $m \leq n$. The process is repeated until the total budget, $b_{total} = \sum_{t=1}^{T} {b_t}$, is exhausted. Henceforth, we will use the same notations where applicable unless otherwise stated.

\subsection{Core-set Approach}
\label{coreset_theory}
This section follows \citep{sener2017active} very closely, with some important alterations to adapt to regression problems. Core-set selection aims to train a model using only a subset of data and achieve competitive results compared to using the entire dataset \citep{sener2017active}. Consider a large pool of data sampled $i.i.d.$ over the space $\mathcal{Z} = \mathcal{X} \times \mathcal{Y}$; at time step $t+1$, by choosing a subset of the data, we seek to minimise the future expected loss:
\begin{equation}
\label{eq:min_expected_loss}
    \min_{\mathbf{s}^{t+1}: |\mathbf{s}^{t+1}| \leq b_{t+1}} E_{\mathbf{x}, \mathbf{y} \sim p_z} \left[ l(\mathbf{x}, \mathbf{y}; A_{\bigcup_{\tau=1}^{t+1}\mathbf{s}^{\tau}}) \right]
\end{equation}{where %$\mathbf{s}^t$ is the labelled data collected in each time step $t$,
$A_{\mathbf{s}}$ is a learning algorithm which outputs parameters $\mathbf{w}$ given labelled data $\mathbf{s}$, and $l(\mathbf{x}, \mathbf{y}; A_\mathbf{s})$ is the loss between ground truth $\mathbf{y}$ and the emulated output given input $\mathbf{x}$ and parameters $\mathbf{w}$.}
% This approach is geometrically motivated, and it turns out to be equivalent to solving the k-Center problem \citep{wolf2011facility} when a CNNs-based architecture is used. We state the theorem here, and give the proof to the theorem in the appendix. We note that this theorem and its proof follow those in \citep{sener2017active}, but with some important modifications to adapt to regression settings.

\autoref{eq:min_expected_loss} can then be upper bounded by a summation of training error, generalisation error and core-set loss. In an expressive model, training and generalisation errors are assumed to be negligible, allowing us to redefine the problem as core-set loss minimisation:
% The authors in \citep{sener2017active} gave an upper bound to \autoref{eq:min_expected_loss} by noting that the training and generalisation errors of a good model are typically negligible (we replicated in \autoref{loss} for completion) and redefined the active learning problem to minimise the following upper bound, termed core-set loss: 
\begin{equation}
    \min_{\mathbf{s}^{t+1}: |\mathbf{s}^{t+1}| \leq b_{t+1}} \Bigl| \frac{1}{n}\sum_{i \in [n]} l(\mathbf{x}_i,\mathbf{y}_i;A_{\bigcup_{\tau=1}^{t+1}\mathbf{s}^{\tau}}) - \frac{1}{|\bigcup_{\tau=1}^{t+1}\mathbf{s}^{\tau}|}\sum_{j \in \bigcup_{\tau=1}^{t+1}\mathbf{s}^{\tau}} l(\mathbf{x}_j,\mathbf{y}_j;A_{\bigcup_{\tau=1}^{t+1}\mathbf{s}^{\tau}}) \Bigr|
\end{equation}

\begin{lemma}
\label{lemma1}
Loss function defined as the $L_{1}$-norm between the true value and the model output of a ReLU-activated convolutional neural network with $\eta_c$ convolutional layers and $\eta_{fc}$ fully-connected layers is $\alpha^{\eta_c+\eta_{fc}}$-Lipschitz continuous.
\end{lemma}
\begin{theorem}
\label{theorem1}
Given $n$ i.i.d. samples drawn from $p_z$ as $\{\mathbf{x}_i, \mathbf{y}_i\}_{i \in [n]}$ and set of points $\mathbf{s}$, if loss function $l(.,\mathbf{y};\mathbf{w})$ is $\lambda^l$-Lipschitz continuous for all $\mathbf{y}, \mathbf{w}$ and bounded by $L$, simulation regression function, denoted by $\eta(\mathbf{x})=p(\mathbf{y}|\mathbf{x})$, is $\lambda^\eta$-Lipschitz for all $\mathbf{y}$, $\mathbf{s}$ is $\delta_s$ cover of $\{\mathbf{x}_i, \mathbf{y}_i\}_{i \in [n]}$, and $l(\mathbf{x}_{\mathbf{s}(j)},\mathbf{y}_{\mathbf{s}(j)};A_{\mathbf{s}})=0$ for all $j\in[m]$, with a probability of at least $1-\gamma$,
\begin{equation*}
     % P \left[ \Bigl| \frac{1}{n}\sum l(\mathbf{x}_i,\mathbf{y}_i;A_{\mathbf{s}}) - \frac{1}{|\mathbf{s}|}\sum l(\mathbf{x}_j,\mathbf{y}_j;A_{\mathbf{s}}) \Bigr| \leq \delta \left [ \lambda^{l} + \lambda^{\eta}LV \right] + \sqrt{\frac{L^2\log(1/\gamma)}{2n}} \right] \geq 1 - \gamma
     \Bigl| \frac{1}{n}\sum_{i \in [n]} l(\mathbf{x}_i,\mathbf{y}_i;A_{\mathbf{s}}) - \frac{1}{|\mathbf{s}|}\sum_{j \in \mathbf{s}} l(\mathbf{x}_j,\mathbf{y}_j;A_{\mathbf{s}}) \Bigr| \leq \delta \left [ \lambda^{l} + \lambda^{\eta}LV \right] + \sqrt{\frac{L^2\log(1/\gamma)}{2n}}
\end{equation*}
\end{theorem}{``$\mathbf{s}$ is $\delta_s$ cover of $\mathbf{s}^*$'' informally means $\mathbf{s}$ are the centers of a set of circles with radii $\delta$, and the circles can cover all $\mathbf{s}^*$. $V$ in \autoref{theorem1} is the hypervolume that depends on the support over the $\mathcal{Y}$ space for each $\mathbf{x}$. We consider scientific simulations %that are noiseless and deterministic with dimension
with very little noise and dimension of $\mathcal{Y} \gg 1$, the term that contains $V$ can practically be ignored, giving a tighter upper bound in practice. It also reveals an important, albeit counter-intuitive insight: for a simulation with very little noise, we get a tighter upper bound of the expected loss when a higher dimension of $\mathcal{Y}$ is used, i.e. more granular outputs are modelled. % If the features $\mathbf{x}$'s are taken from the output of the model, then $\lambda^l=1$.
In essence, with a high probability, the expected future loss is theoretically bounded if an appropriate subset of data is labelled for training; and it turns out, empirically, as we will see in \autoref{experiments}, that the long tail losses are reduced as well.
% Another important observation is that, if the features $\mathbf{x}$'s are taken from the output of the model, then $L$ is a function of $\delta$, and $\lambda^l=1$.
%Another important observation is that the bound is applicable even to a single data-point (see \autoref{theorem_proof}), thus giving a theoretical evidence as to why the long tail losses are reduced.
}
% in this case (or in general?) L is a function of delta? does this help in proving the loss bound?

\subsection{Cold-start and Warm-start Optimisations}
Conventionally, in each active learning iteration, the deep-learning model is trained from scratch, with each of the trainable parameters reinitialised randomly \citep{sener2017active, ash2019deep}, a technique known as cold-start optimisation. It is, however, computationally preferred to warm start the optimisation, i.e. continue training with the learned parameters from the previous iteration. In practice, warm-starting seems to generalise poorly to test data, but it has been shown that this pathology could be overcome in several important situations using a simple shrink-and-perturb trick \citep{ash2019warm}, where the learned parameters are shrunk by a certain percentage and Gaussian noise is added to them, as captured in \autoref{eq:sp}. We leverage this trick in conjunction with the core-set approach in our case studies to obtain the best possible model given wall-clock time constraint.
\begin{equation}
\label{eq:sp}
    \theta_{i}^t \longleftarrow \lambda\theta_{i}^{t-1} + p^t, \text{ where } p^t \sim \mathcal{N}(0,\,\sigma^2) \text{ and } 0 \leq \lambda \leq 1
\end{equation}{}%where $p^t \sim \mathcal{N}(0,\,\sigma^2)$ and $0 \leq \lambda \leq 1$}
\section{Experiments and Results}
\label{experiments}
\iffalse
\begin{algorithm}
\caption{Active Learning}\label{alg:cap}
\begin{algorithmic}
\Require $n \geq 0$
\Ensure $y = x^n$
\State $y \gets 1$
\State $X \gets x$
\State $N \gets n$
\While{$N \neq 0$}
\If{$N$ is even}
    \State $X \gets X \times X$
    \State $N \gets \frac{N}{2}$  \Comment{This is a comment}
\ElsIf{$N$ is odd}
    \State $y \gets y \times X$
    \State $N \gets N - 1$
\EndIf
\EndWhile
\end{algorithmic}
\end{algorithm}
\fi

\begin{table}
  \caption{Active learning details for emulations in XES and Halo}
  \label{experiment_details}
  \centering
  \begin{tabular}{lll}
    \toprule
    %\multicolumn{3}{c}{Part}                   \\
    %\cmidrule(r){1-2}
         & XES     & Halo \\
    \midrule
    $n$, initial unlabelled sample pool size         & $10,000$ & $50,000$ \\
    $b_t$, budget at time step $t \in \{1, \dots, T\}$      & $1,500$  & $2,000$  \\
    $T$, total number of time steps        & $5$      & $5$     \\
    Validation data size  & $1,000$  & $3,000$ \\
    Test data size   & $3,000$  & $87,000$ \\
    \bottomrule
  \end{tabular}
\end{table} 

\begin{figure}[htp]
    \centering
        \subfloat[Halo]{\includegraphics[width=.5\textwidth]{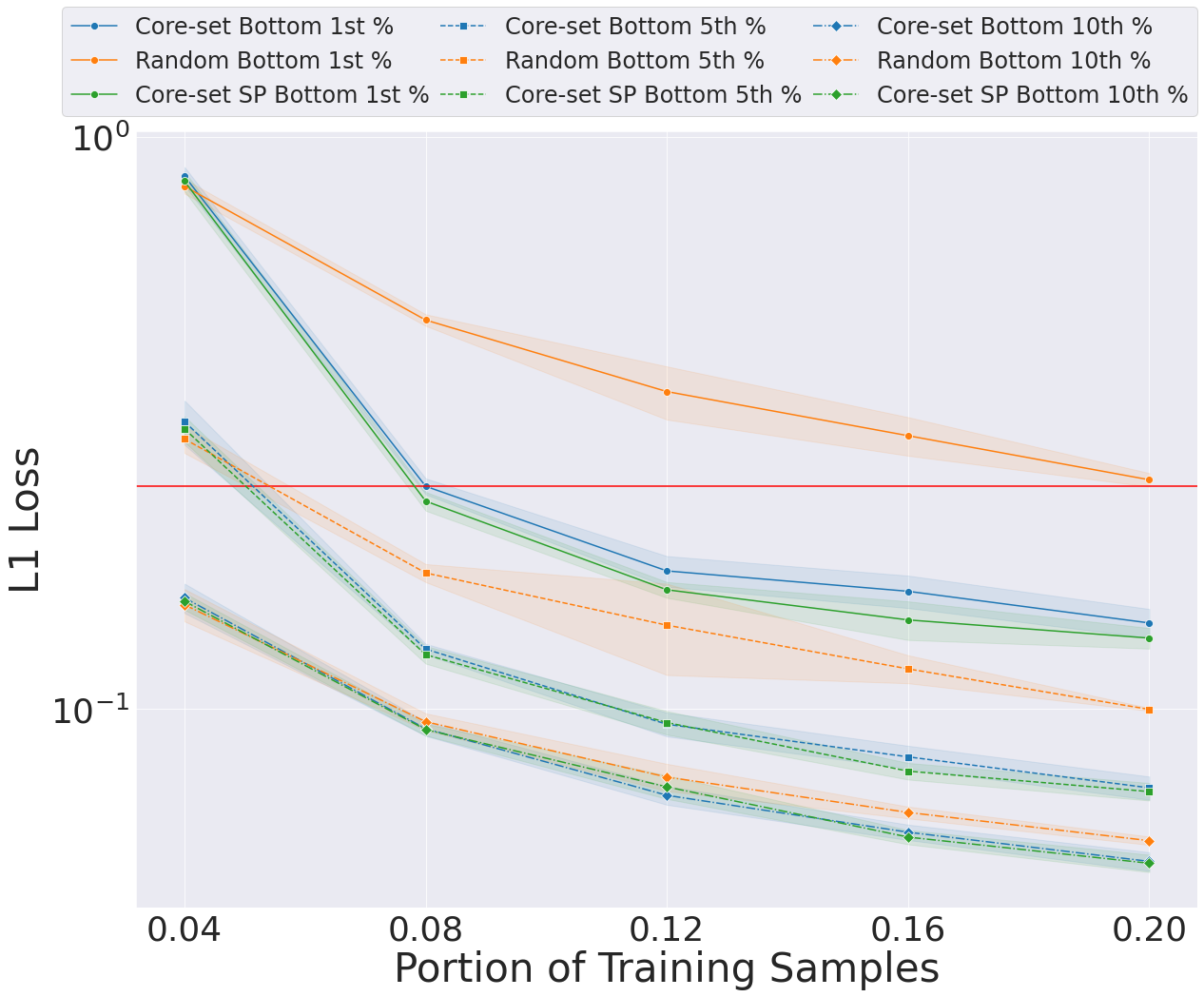}}\hfill
        \subfloat[XES]{\includegraphics[width=.5\textwidth]{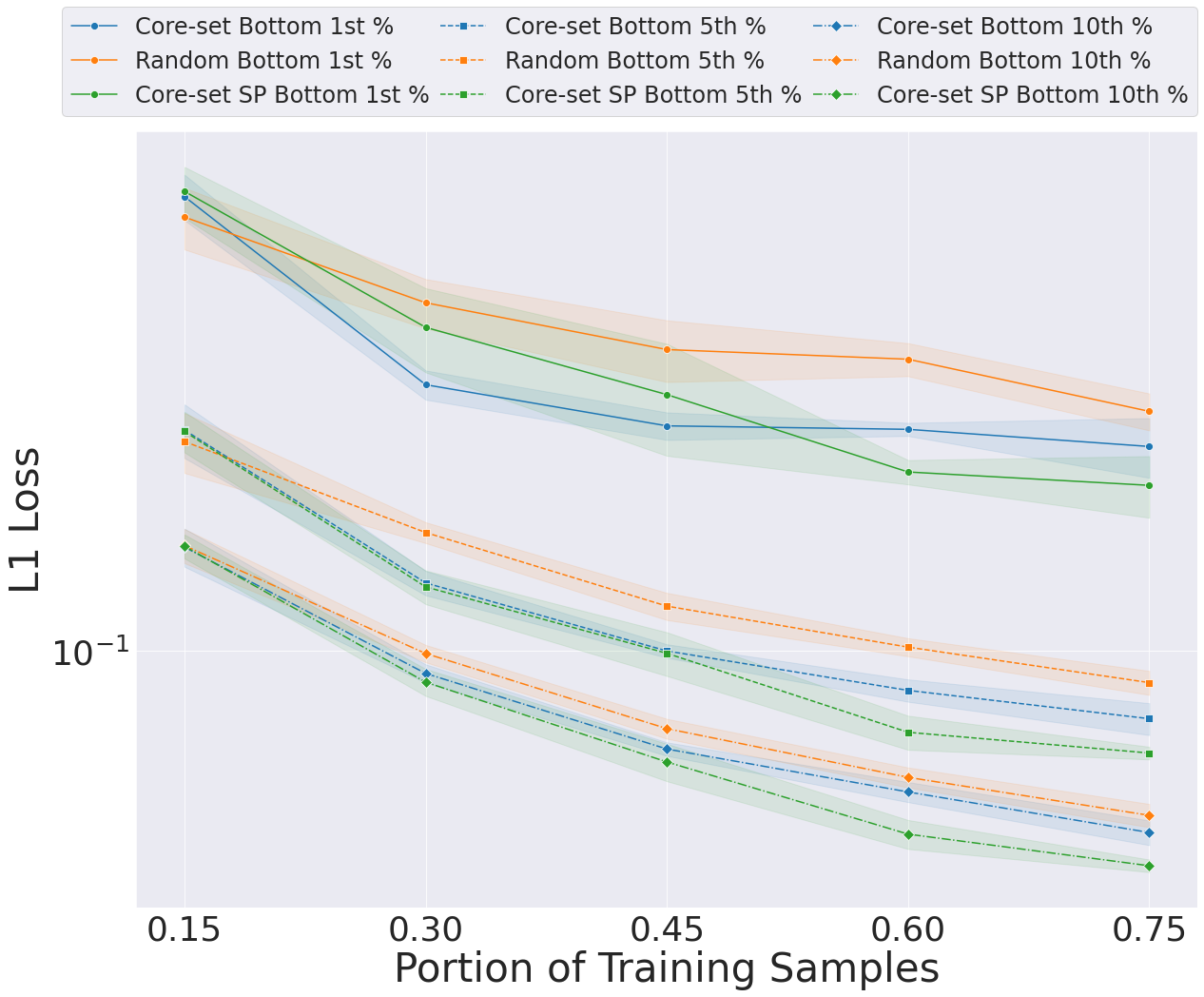}}
    \caption{Comparison between random sampling and core-set selection with Halo emulator and XES emulator. SP indicates shrink-and-perturb trick is used in the training, otherwise the optimisation is cold started. $\%$ represents percentile. Red horizontal line in (a) shows a $60\%$ reduction in labelled data required with core-set selection.}
    \label{fig:bottom_loss}
%\end{figure}

%\begin{figure}[htp]
    \centering
        \subfloat[Random]{\includegraphics[width=.333\textwidth]{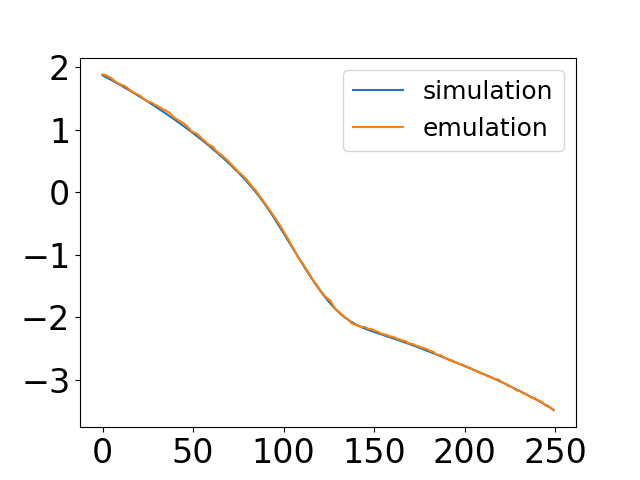}}\hfill
        \subfloat[Core-set]{\includegraphics[width=.333\textwidth]{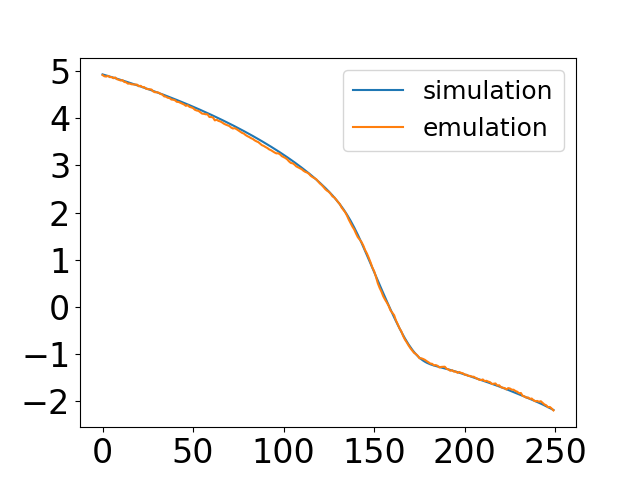}}\hfill
        \subfloat[Core-set SP]{\includegraphics[width=.333\textwidth]{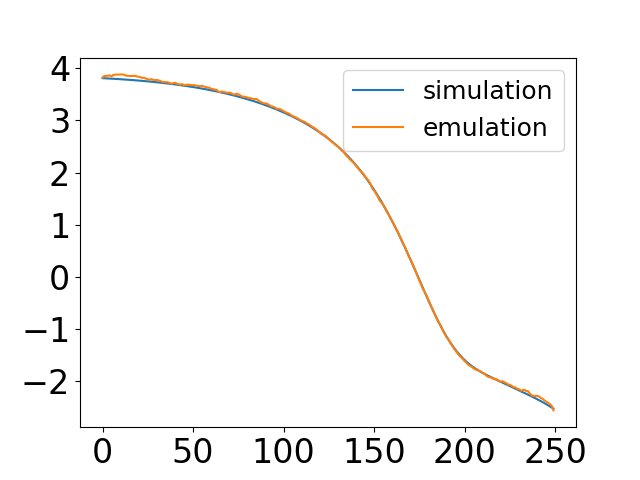}}
    \caption{Median Halo emulations using different sampling methods and optimisation schemes.} %From left to right: random sampling, core-set selection, core-set selection with shrink-and-perturb trick.}
    \label{fig:halo_median}
%\end{figure}

%\begin{figure}[htp]
    \centering
        \subfloat[Random]{\includegraphics[width=.333\textwidth]{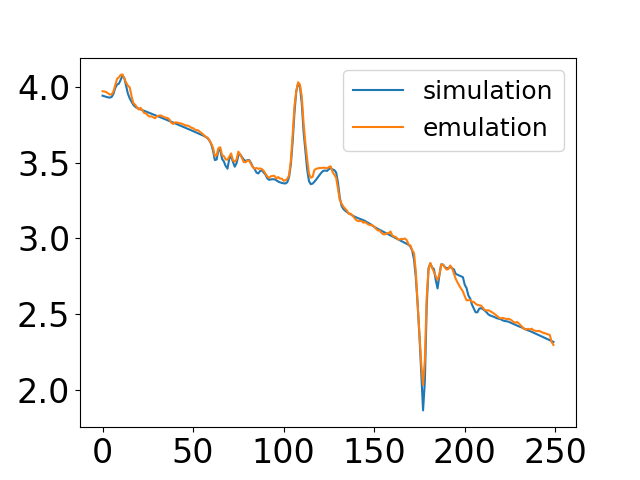}}\hfill
        \subfloat[Core-set]{\includegraphics[width=.333\textwidth]{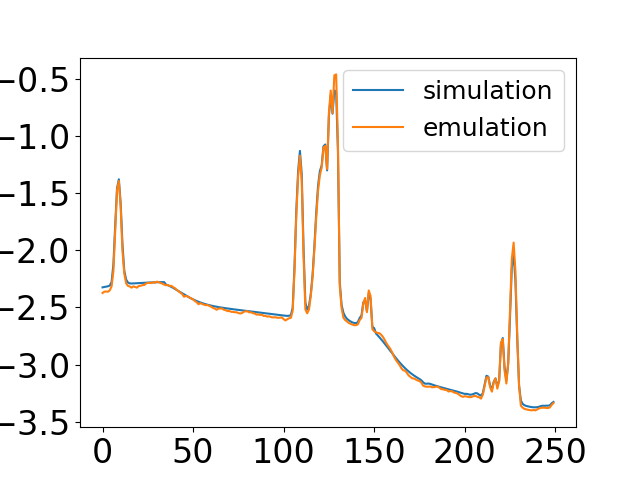}}\hfill
        \subfloat[Core-set SP]{\includegraphics[width=.333\textwidth]{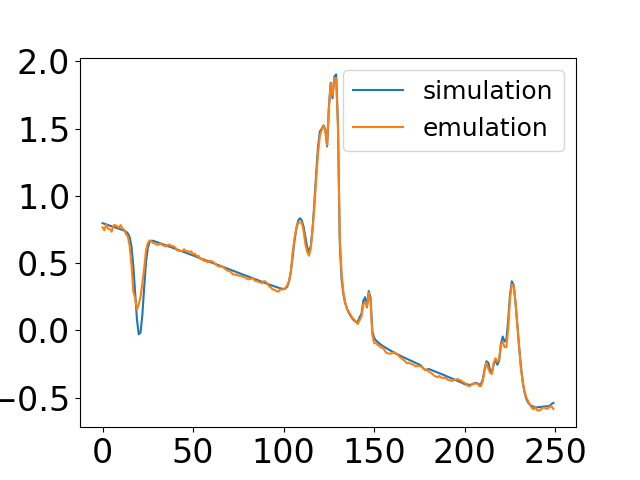}}
    \caption{Median XES emulations using different sampling methods and optimisation schemes.} %From left to right: random sampling, core-set selection, core-set selection with shrink-and-perturb trick.}
    \label{fig:xes_median}
%\end{figure}

%\begin{figure}[htp]
    \centering
        \subfloat[Random]{\includegraphics[width=.333\textwidth]{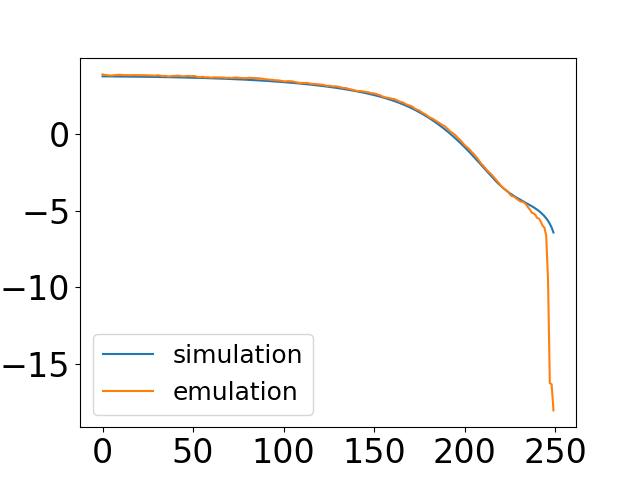}}\hfill
        \subfloat[Core-set]{\includegraphics[width=.333\textwidth]{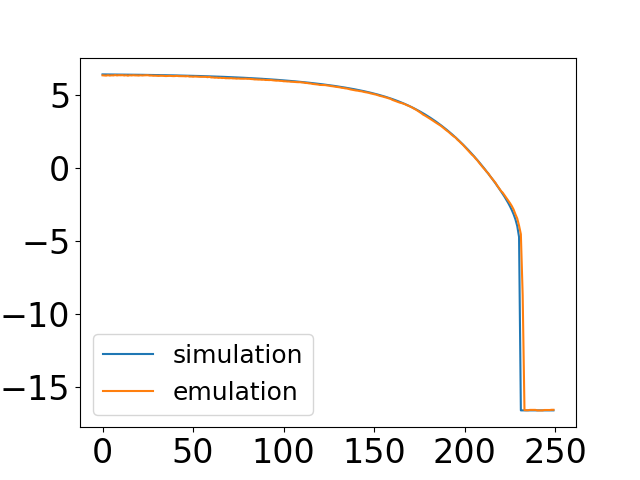}}\hfill
        \subfloat[Core-set SP]{\includegraphics[width=.333\textwidth]{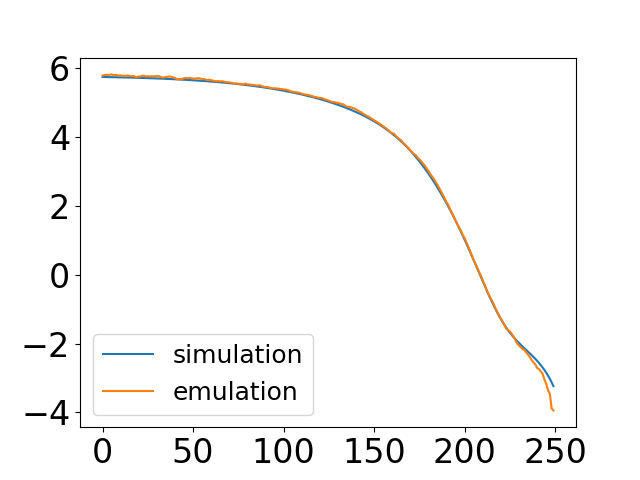}}
    \caption{Bottom first percentile of Halo emulations using different sampling methods and optimisation schemes.} %From left to right: random sampling, core-set selection, core-set selection with shrink-and-perturb trick.}
    \label{fig:halo_bottom}
\end{figure}

A deterministic, supervised CNNs-based model architecture \citep{kasim2021building, he2016deep} was used for all experiments. We trained the models with $L_{1}$-loss function and Adam \citep{kingma2014adam} optimiser with a minibatch size of $32$ and learning rate $0.001$, and judged their performance by the average $L_{1}$ losses of the bottom 1\textsuperscript{st}, 5\textsuperscript{th} and 10\textsuperscript{th} percentiles. We also plotted the median predictions in \autoref{fig:halo_median} and \autoref{fig:xes_median} to show that the emulation quality for the better half of the test cases does not vary much with the sampling method. For core-set selection, we solved the k-center problem at virtually no overhead costs using KeOps library \citep{JMLR:v22:20-275}. Model outputs were used as the features for core-set, effectively bounding the greatest L1 loss among the samples. Other important hyper-parameters are noted in \autoref{experiment_details}. All experiments were run with 4 NVIDIA Tesla T4 GPUs on Google Cloud Platform.

From \autoref{fig:bottom_loss}, it is clear that core-set selection outperforms random sampling consistently in reducing the long tail $L_{1}$ losses in all percentiles investigated. Moreover, we needed significantly less data when using core-set selection to achieve similar performance as their random sampling counterpart. Notably, to improve the bottom first percentile of the Halo emulation to a competitive level, a whopping $60\%$ reduction in costs was seen (see \autoref{fig:bottom_loss}) using core-set selection. This is highlighted in \autoref{fig:halo_bottom} where the emulator failed to predict part of the signal when it was trained with $10,000$ randomly sampled data, but successfully emulated similar signals using data from core-set.

Furthermore, we recorded marginal, but important improvement in the long tail $L_{1}$ losses when the shrink-and-perturb trick is used in conjunction with the core-set approach, with $\sigma=0.1$ and $\lambda=0.5$. 
%. In both case studies, we set $\sigma=0.1$ and $\lambda=0.5$. 
% This finding is interesting as the learnt hypothesis of the model is technically not preserved upon shrinking the parameters in a regression problem. We leave the explanation as a future research direction. 

% Interestingly, the median predictions (\autoref{fig:halo_median} and \autoref{fig:xes_median}) using random sampling record the smallest losses among all sampling methods, even though the mean losses in all percentiles investigated are consistently higher than the other methods. 

% This means that by using random sampling, the model tries to minimise the losses of the majority of the training data, at the expense of a few outliers. This is unfavourable since we seek to reduce both the average loss and the right long tail of the loss distribution concurrently.

% To test if the shrink-and-perturb trick works in regression, we shrank the model parameters by $50\%$, and added random Gaussian noise centered at $0$ with a standard deviation of $0.1$ to each trainable parameter. Our results suggest that with core-set data, models with weights initialised in each iteration using the shrink-and-perturb trick perform comparably with, and in some cases even better than, those with weights reinitialised from scratch.

\section{Conclusions}
We investigated how a diversity-based active learning approach, core-set selection, affects the long tail losses in scientific emulations, and how the active learning iterations could be warm started to improve the model performance and save costs. Our experiments show promising results: the long tail losses were significantly reduced without compromising the quality of the better emulations.

%Most of the existing papers in the active learning area focus on classification problems, with the algorithms derived based on the premise that a sigmoid activation function is used right before the output layer, which is a clear violation in many scientific emulations. We hope that this study can serve as a starting point for new research into more general active learning approaches that are theoretically sound for both classification and regression problems. 

\section{Broader Impact}
\label{broader_impact}
On the negative side, designing a good, explainable model architecture to replace the theoretically-grounded scientific simulations could be time-consuming, and a performant model is not necessarily explicable. Besides, modern deep learning is over-dependent on supervised learning, where the performance is in turn reliant on the availability of labelled data. 

Our work partially solves the data problem. Using actively sampled, good quality data could improve the general performance of supervised emulators with less data required and thus accelerate scientific research. Moreover, replacing simulators with neural networks trained with actively selected data would significantly reduce computational costs and the adverse environmental impacts that come with heavy computations. 
\FloatBarrier

\clearpage

\bibliographystyle{unsrtnat}
\setcitestyle{numbers,square,super}
\bibliography{ref}

\begin{thebibliography}{16}
\providecommand{\natexlab}[1]{#1}
\providecommand{\url}[1]{\texttt{#1}}
\expandafter\ifx\csname urlstyle\endcsname\relax
  \providecommand{\doi}[1]{doi: #1}\else
  \providecommand{\doi}{doi: \begingroup \urlstyle{rm}\Url}\fi

\bibitem[Brockherde et~al.(2017)Brockherde, Vogt, Li, Tuckerman, Burke, and
  M{\"u}ller]{brockherde2017bypassing}
Felix Brockherde, Leslie Vogt, Li~Li, Mark~E Tuckerman, Kieron Burke, and
  Klaus-Robert M{\"u}ller.
\newblock Bypassing the kohn-sham equations with machine learning.
\newblock \emph{Nature communications}, 8\penalty0 (1):\penalty0 1--10, 2017.

\bibitem[Kwan et~al.(2015)Kwan, Heitmann, Habib, Padmanabhan, Lawrence, Finkel,
  Frontiere, and Pope]{kwan2015cosmic}
Juliana Kwan, Katrin Heitmann, Salman Habib, Nikhil Padmanabhan, Earl Lawrence,
  Hal Finkel, Nicholas Frontiere, and Adrian Pope.
\newblock Cosmic emulation: fast predictions for the galaxy power spectrum.
\newblock \emph{The Astrophysical Journal}, 810\penalty0 (1):\penalty0 35,
  2015.

\bibitem[Peterson et~al.(2017)Peterson, Humbird, Field, Brandon, Langer, Nora,
  Spears, and Springer]{peterson2017zonal}
Jayson~L Peterson, KD~Humbird, John~E Field, Scott~T Brandon, Steve~H Langer,
  Ryan~C Nora, Brian~K Spears, and PT~Springer.
\newblock Zonal flow generation in inertial confinement fusion implosions.
\newblock \emph{Physics of Plasmas}, 24\penalty0 (3):\penalty0 032702, 2017.

\bibitem[Rupp et~al.(2012)Rupp, Tkatchenko, M{\"u}ller, and
  Von~Lilienfeld]{rupp2012fast}
Matthias Rupp, Alexandre Tkatchenko, Klaus-Robert M{\"u}ller, and O~Anatole
  Von~Lilienfeld.
\newblock Fast and accurate modeling of molecular atomization energies with
  machine learning.
\newblock \emph{Physical review letters}, 108\penalty0 (5):\penalty0 058301,
  2012.

\bibitem[Kasim et~al.(2021)Kasim, Watson-Parris, Deaconu, Oliver, Hatfield,
  Froula, Gregori, Jarvis, Khatiwala, Korenaga, et~al.]{kasim2021building}
Muhammad~Firmansyah Kasim, D~Watson-Parris, L~Deaconu, S~Oliver, P~Hatfield,
  Dustin~H Froula, Gianluca Gregori, M~Jarvis, S~Khatiwala, J~Korenaga, et~al.
\newblock Building high accuracy emulators for scientific simulations with deep
  neural architecture search.
\newblock \emph{Machine Learning: Science and Technology}, 2021.

\bibitem[Sener and Savarese(2017)]{sener2017active}
Ozan Sener and Silvio Savarese.
\newblock Active learning for convolutional neural networks: A core-set
  approach.
\newblock \emph{arXiv preprint arXiv:1708.00489}, 2017.

\bibitem[Gal et~al.(2017)Gal, Islam, and Ghahramani]{gal2017deep}
Yarin Gal, Riashat Islam, and Zoubin Ghahramani.
\newblock Deep bayesian active learning with image data.
\newblock In \emph{International Conference on Machine Learning}, pages
  1183--1192. PMLR, 2017.

\bibitem[Ash et~al.(2019)Ash, Zhang, Krishnamurthy, Langford, and
  Agarwal]{ash2019deep}
Jordan~T Ash, Chicheng Zhang, Akshay Krishnamurthy, John Langford, and Alekh
  Agarwal.
\newblock Deep batch active learning by diverse, uncertain gradient lower
  bounds.
\newblock \emph{arXiv preprint arXiv:1906.03671}, 2019.

\bibitem[Kirsch et~al.(2019)Kirsch, Van~Amersfoort, and
  Gal]{kirsch2019batchbald}
Andreas Kirsch, Joost Van~Amersfoort, and Yarin Gal.
\newblock Batchbald: Efficient and diverse batch acquisition for deep bayesian
  active learning.
\newblock \emph{Advances in neural information processing systems},
  32:\penalty0 7026--7037, 2019.

\bibitem[Regan et~al.(2013)Regan, Epstein, Hammel, Suter, Scott, Barrios,
  Bradley, Callahan, Cerjan, Collins, et~al.]{regan2013hot}
SP~Regan, R~Epstein, BA~Hammel, LJ~Suter, HA~Scott, MA~Barrios, DK~Bradley,
  DA~Callahan, C~Cerjan, GW~Collins, et~al.
\newblock Hot-spot mix in ignition-scale inertial confinement fusion targets.
\newblock \emph{Physical review letters}, 111\penalty0 (4):\penalty0 045001,
  2013.

\bibitem[Ciricosta et~al.(2017)Ciricosta, Scott, Durey, Hammel, Epstein,
  Preston, Regan, Vinko, Woolsey, and Wark]{ciricosta2017simultaneous}
Orlando Ciricosta, H~Scott, P~Durey, BA~Hammel, R~Epstein, TR~Preston,
  SP~Regan, SM~Vinko, NC~Woolsey, and JS~Wark.
\newblock Simultaneous diagnosis of radial profiles and mix in nif
  ignition-scale implosions via x-ray spectroscopy.
\newblock \emph{Physics of Plasmas}, 24\penalty0 (11):\penalty0 112703, 2017.

\bibitem[Wake et~al.(2011)Wake, Whitaker, Labb{\'e}, Van~Dokkum, Franx, Quadri,
  Brammer, Kriek, Lundgren, Marchesini, et~al.]{wake2011galaxy}
David~A Wake, Katherine~E Whitaker, Ivo Labb{\'e}, Pieter~G Van~Dokkum, Marijn
  Franx, Ryan Quadri, Gabriel Brammer, Mariska Kriek, Britt~F Lundgren, Danilo
  Marchesini, et~al.
\newblock Galaxy clustering in the newfirm medium band survey: the relationship
  between stellar mass and dark matter halo mass at 1< z< 2.
\newblock \emph{The Astrophysical Journal}, 728\penalty0 (1):\penalty0 46,
  2011.

\bibitem[Ash and Adams(2019)]{ash2019warm}
Jordan~T Ash and Ryan~P Adams.
\newblock On warm-starting neural network training.
\newblock \emph{arXiv preprint arXiv:1910.08475}, 2019.

\bibitem[Charlier et~al.(2021)Charlier, Feydy, Glaunès, Collin, and
  Durif]{JMLR:v22:20-275}
Benjamin Charlier, Jean Feydy, Joan~Alexis Glaunès, François-David Collin,
  and Ghislain Durif.
\newblock Kernel operations on the gpu, with autodiff, without memory
  overflows.
\newblock \emph{Journal of Machine Learning Research}, 22\penalty0
  (74):\penalty0 1--6, 2021.
\newblock URL \url{http://jmlr.org/papers/v22/20-275.html}.

\bibitem[He et~al.(2016)He, Zhang, Ren, and Sun]{he2016deep}
Kaiming He, Xiangyu Zhang, Shaoqing Ren, and Jian Sun.
\newblock Deep residual learning for image recognition.
\newblock In \emph{Proceedings of the IEEE conference on computer vision and
  pattern recognition}, pages 770--778, 2016.

\bibitem[Kingma and Ba(2014)]{kingma2014adam}
Diederik~P Kingma and Jimmy Ba.
\newblock Adam: A method for stochastic optimization.
\newblock \emph{arXiv preprint arXiv:1412.6980}, 2014.

\end{thebibliography}

\clearpage
\clearpage
\begin{appendices}
\section{Future Expected Loss}
\label{loss}
The future expected loss can be upper bounded by the summation of three terms, namely generalisation error, training error and core-set loss. In an expressive model such as CNNs, the generalisation and training errors can be assumed to be negligible, thus simplifying the upper bound to just the core-set loss.
\begin{align*}
    E_{\mathbf{x}, \mathbf{y}\sim p_{z}} \left[ l(\mathbf{x},\mathbf{y};A_{\mathbf{s}}) \right]
    &\leq \begin{aligned}[t]
                &\underbrace{\Bigl| E_{\mathbf{x}, \mathbf{y}\sim p_{z}} \left[ l(\mathbf{x},\mathbf{y};A_{\mathbf{s}}) \right] - \frac{1}{n}  \sum_{i=1}^{n} l(\mathbf{x}_i,\mathbf{y}_i;A_{\mathbf{s}}) \Bigr|}_{\text{Generalisation Error}} \\
                &+\underbrace{\frac{1}{|\mathbf{s}|}\sum_{j\in\mathbf{s}} l(\mathbf{x}_j,\mathbf{y}_j;A_{\mathbf{s}})}_{\text{Training Error}}\\
                &+\underbrace{\Bigl| \frac{1}{n}  \sum_{i=1}^{n} l(\mathbf{x}_i,\mathbf{y}_i;A_{\mathbf{s}}) - \frac{1}{|\mathbf{s}|}\sum_{j\in\mathbf{s}} l(\mathbf{x}_j,\mathbf{y}_j;A_{\mathbf{s}}) \Bigr|}_{\text{Core-set Loss}}
          \end{aligned} 
\end{align*}{}
\section{Proof for Lemma 1}
\label{lemma_proof}
Consider two vectors $\mathbf{x}^{d}$ and $\mathbf{\Tilde{x}}^{d}$ that correspond to inputs $\mathbf{x}$ and $\mathbf{\Tilde{x}}$ at layer $d$ of a neural net respectively. First, for any convolutional or fully connected layer, for each node $j$ in layer $d$, we can write $\mathbf{x}_{j}^{d} = \sum_{i} w^{d}_{i,j}\mathbf{x}_i^{d-1}$. Now, assuming $\sum_{i}|w_{i,j}| \leq \alpha$ for all $i, \: j, \: d$, we can write, for any convolutional or fully connected layer:

\begin{equation}
\label{eq:lipschitz_linear}
    ||\mathbf{x}^d - \mathbf{\Tilde{x}}^d||_{q} \leq \alpha ||\mathbf{x}^{d-1} - \mathbf{\Tilde{x}}^{d-1}||_{q}
\end{equation}{for any real $q \geq 1$.}
% should hold for any q \gg 1, check again

For ReLU, we note that $|a-b| \geq |\max{(0, a)}-\max{(0,b)}|$, and as such we can state:
\begin{equation}
\label{eq:lipschitz_relu}
    ||\mathbf{x}^r - \mathbf{\Tilde{x}}^r||_{q} \leq ||\mathbf{x}^{r-1} - \mathbf{\Tilde{x}}^{r-1}||_{q}
\end{equation}{where the superscripts $r-1$ and $r$ indicate before and after the ReLU activation function respectively.}

Combining \autoref{eq:lipschitz_linear} and \autoref{eq:lipschitz_relu}, for ReLU-activated neural networks with $\eta_c$ convolutional layers and $\eta_{fc}$ fully connected layers, we get
\begin{equation*}
    ||CNN(\mathbf{x};\mathbf{w}) - CNN(\mathbf{\Tilde{x}};\mathbf{w})||_q \leq \alpha^{\eta_c + \eta_{fc}} ||\mathbf{x} - \mathbf{\Tilde{x}}||_q
\end{equation*}

Finally, using reverse triangle inequality, we obtain
\begin{equation*}
    |l(\mathbf{x}, \mathbf{y}; \mathbf{w}) - l(\mathbf{\Tilde{x}}, \mathbf{y}; \mathbf{w})|
    = |||CNN(\mathbf{x};\mathbf{w})-\mathbf{y}||_q - ||CNN(\mathbf{\Tilde{x}};\mathbf{w})-\mathbf{y}||_q|
    \leq ||CNN(\mathbf{x};\mathbf{w}) - CNN(\mathbf{\Tilde{x}};\mathbf{w})||_q 
\end{equation*}{hence finishing the proof.}

\section{Proof for Theorem 1}
\label{theorem_proof}
We note that for two probability density functions $p$ and $p^*$ defined over the space $\mathcal{Y}$, the following is always true:

\begin{equation}
\label{eq:pdf_inequality}
    \int p \,dy \leq \int p^* \,dy + \int | p - p^*| \,dy
\end{equation}{}
Then, we bound $E_{\mathbf{y}\sim \eta({\mathbf{x}_{i}})} \left[ l(\mathbf{x}_i, \mathbf{y};A_{\mathbf{s}}) \right]$ by using \autoref{eq:pdf_inequality}:

\begin{align*}
    E_{\mathbf{y}\sim \eta({\mathbf{x}_i})} \left[ l(\mathbf{x}_i,\mathbf{y};A_{\mathbf{s}}) \right]
    &= \int p_{\mathbf{y}\sim\eta(\mathbf{x}_i)}(\mathbf{y})\left[ l(\mathbf{x}_i,\mathbf{y};A_{\mathbf{s}}) \right] \,d\mathbf{y} \\
    &\leq \begin{aligned}[t] 
            &\int p_{\mathbf{y}\sim\eta(\mathbf{x}_j)}(\mathbf{y})\left[ l(\mathbf{x}_i,\mathbf{y};A_{\mathbf{s}}) \right] \,d\mathbf{y} \\
            &+ \int |p_{\mathbf{y}\sim\eta(\mathbf{x}_i)}(\mathbf{y})-p_{\mathbf{y}\sim\eta(\mathbf{x}_j)}(\mathbf{y})|\left[ l(\mathbf{x}_i,\mathbf{y};A_{\mathbf{s}}) \right] \,d\mathbf{y}
          \end{aligned} \\
    &\leq \int p_{\mathbf{y}\sim\eta(\mathbf{x}_j)}(\mathbf{y})\left[ l(\mathbf{x}_i,\mathbf{y};A_{\mathbf{s}}) \right] \,d\mathbf{y} + \delta\lambda^{\eta}L\int d\mathbf{y}\\
    &=\int p_{\mathbf{y}\sim\eta(\mathbf{x}_j)}(\mathbf{y})\left[ l(\mathbf{x}_i,\mathbf{y};A_{\mathbf{s}}) \right] \,d\mathbf{y} + \delta\lambda^{\eta}LV
\end{align*}{where $\mathbf{x}_j$ for $j \in [m]$ are labelled training data. Then, recall the assumption that the trained networks would fit perfectly to the training samples and using Lemma 1, we get
}
\begin{align*}
    \int p_{\mathbf{y}\sim\eta(\mathbf{x}_j)}(\mathbf{y})\left[ l(\mathbf{x}_i,\mathbf{y};A_{\mathbf{s}}) \right] \,d\mathbf{y} 
    &= \begin{aligned}[t] 
            &\int p_{\mathbf{y}\sim\eta(\mathbf{x}_j)}(\mathbf{y})\left[ l(\mathbf{x}_i,\mathbf{y};A_{\mathbf{s}}) - l(\mathbf{x_{j}},\mathbf{y};A_{\mathbf{s}}) \right] \,d\mathbf{y} \\
            &+\int p_{\mathbf{y}\sim\eta(\mathbf{x}_j)}(\mathbf{y})\left[ l(\mathbf{x_{j}},\mathbf{y};A_{\mathbf{s}}) \right] \,d\mathbf{y}
       \end{aligned} \\
    &\leq \delta\lambda^{l}
\end{align*}{which gives us}
\begin{equation*}
    E_{\mathbf{y}\sim \eta({\mathbf{x}_i})} \left[ l(\mathbf{x}_i,\mathbf{y};A_{\mathbf{s}}) \right] \leq \delta \left [ \lambda^{l} + \lambda^{\eta}LV \right] 
\end{equation*}{Finally, using Hoeffding's Bound, %\citep{hoeffding1994probability}
we conclude that with a probability of at least $1-\gamma$, we get an upper bound for the core-set loss,}
\begin{equation*}
  \Bigl| \frac{1}{n}\sum_{i \in [n]} l(\mathbf{x}_i,\mathbf{y}_i;A_{\mathbf{s}}) - \frac{1}{|\mathbf{s}|}\sum_{j \in \mathbf{s}} l(\mathbf{x}_j,\mathbf{y}_j;A_{\mathbf{s}}) \Bigr| \leq \delta \left [ \lambda^{l} + \lambda^{\eta}LV \right] + \sqrt{\frac{L^2\log(1/\gamma)}{2n}}
\end{equation*}
\end{appendices}
\clearpage

\end{document}